%% file: lstm-recursion-paper.tex
\documentclass{article}
\usepackage{nips15submit_e,times}
\usepackage{latexsym}

\usepackage{url}
\usepackage[leqno, fleqn]{amsmath}
\usepackage{amssymb}
\usepackage{qtree}
\usepackage{graphicx}
\usepackage{booktabs}
\usepackage{colortbl}
\usepackage{caption}
\usepackage{subcaption}
\usepackage{xcolor}
\usepackage{color}
\usepackage{tikz}
\usepackage{todonotes}
\usepackage[numbers]{natbib}

\newcount\colveccount
\newcommand*\colvec[1]{
        \global\colveccount#1
        \begin{bmatrix}
        \colvecnext
}
\def\colvecnext#1{
        #1
        \global\advance\colveccount-1
        \ifnum\colveccount>0
                \\
                \expandafter\colvecnext
        \else
                \end{bmatrix}
        \fi
}

\newcommand{\nateq}{\equiv}
\newcommand{\natind}{\mathbin{\#}}
\newcommand{\natneg}{\mathbin{^{\wedge}}}
\newcommand{\natfor}{\sqsubset}
\newcommand{\natrev}{\sqsupset}
\newcommand{\natalt}{\mathbin{|}}
\newcommand{\natcov}{\mathbin{\smallsmile}}

\newcommand{\qq}{\hspace{0.1em}}
\newcommand{\plneg}{\mathop{\textit{not}\qq}}
\newcommand{\pland}{\mathbin{\qq\textit{and}\qq}}
\newcommand{\plor}{\mathbin{\qq\textit{or}\qq}}

\newlength{\howlong}

\usepackage{stmaryrd}

\usepackage{gb4e}
\noautomath

\def\ii#1{\textit{#1}}
\newcommand{\word}[1]{\emph{#1}}
\newcommand{\newcite}[1]{\cite{#1}}

\title{Tree-Structured Composition in Neural Networks\\without Tree-Structured Architectures}

\author{
Samuel R.\ Bowman, Christopher D. Manning, and Christopher Potts\\
Stanford University\\
Stanford, CA 94305-2150\\
\texttt{\{sbowman, manning, cgpotts\}@stanford.edu} \\
}

\date{}

\makeatletter
\newcommand{\@BIBLABEL}{\@emptybiblabel}
\newcommand{\@emptybiblabel}[1]{}
\definecolor{black}{rgb}{0,0,0}
\makeatother
\usepackage[breaklinks, draft, colorlinks, linkcolor=black, urlcolor=black, citecolor=black]{hyperref}

\nipsfinalcopy

\begin{document}
\maketitle

\input{abstract.tex}
\input{intro.tex}
\input{data.tex}

\input{methods.tex}

\input{discussion.tex}



\subsubsection*{Acknowledgments}

We gratefully acknowledge %
a Google Faculty Research Award, %
a gift from Bloomberg L.P., 
and support from
the Defense Advanced Research Projects Agency (DARPA) Deep Exploration and Filtering of Text (DEFT) Program under Air Force Research Laboratory (AFRL) contract no.~FA8750-13-2-0040,
the National Science Foundation under grant no.~IIS 1159679, and %
the Department of the Navy, Office of Naval Research, under grant no.~N00014-13-1-0287.
Any opinions, findings, and conclusions or recommendations expressed in this material are those of the authors and do not necessarily reflect the views of 
Google, 
Bloomberg L.P.,
DARPA,
AFRL,
NSF, 
ONR, or 
the US government.

\bibliographystyle{unsrtnat}
\bibliography{MLSemantics} 

\end{document}

%% file: abstract.tex
\begin{abstract} 
Tree-structured neural networks encode a particular tree geometry for a sentence in the network design. However, these models have at best only slightly outperformed simpler sequence-based models. We hypothesize that neural sequence models like LSTMs are in fact able to discover and implicitly use recursive compositional structure, at least for tasks with clear cues to that structure in the data. We demonstrate this possibility using an artificial data task for which recursive compositional structure is crucial, and find an LSTM-based sequence model can indeed learn to exploit the underlying tree structure. However, its performance consistently lags behind that of tree models, even on large training sets, suggesting that tree-structured models are more effective at exploiting recursive structure.
\end{abstract} 

%% file: intro.tex
\section{Introduction}\label{sec:intro}

Neural networks that encode sentences as real-valued vectors have been successfully used in a wide array of NLP tasks, including translation \cite{sutskever2014sequence}, parsing \cite{dyer2015transition}, and sentiment analysis \cite{tai2015improved}. 
These models are generally either 
sequence models based on recurrent neural networks, which build representations incrementally from left to right \cite{elman1990finding,sutskever2014sequence}, or tree-structured models based on \ii{recursive} neural networks, which build representations incrementally 
according to the hierarchical structure of linguistic phrases \cite{goller1996learning,socher2011semi}. 

While both model classes perform
well on many tasks, and both are under active development,
tree models are often presented as the more principled choice, since they align with standard linguistic assumptions about constituent structure and the compositional derivation of complex meanings.
Nevertheless,
tree models have not shown the kinds of dramatic performance improvements over sequence models that their billing would lead one to expect: head-to-head comparisons with sequence models show either modest improvements \cite{tai2015improved} or none at all \cite{li2015tree}. 

We propose a possible explanation for these results: standard sequence models can learn to
exploit recursive syntactic structure in generating representations of sentence meaning, thereby 
learning to use the 
structure that tree models are explicitly designed around. This requires that
sequence models be able to identify syntactic structure in natural language. We believe this is plausible on the basis of other recent research \cite{vinyals2014grammar,Karpathy2015vaurn}.
In this paper, we  evaluate whether LSTM sequence models are able to use such structure to guide interpretation, 
focusing on cases where syntactic structure is clearly indicated in the data.

We compare standard tree and sequence models on their handling of recursive structure by training the models on sentences whose length and recursion depth are limited, and then testing them on longer and more complex sentences, such that only models that exploit the recursive structure will be able to generalize in a way that yields correct interpretations for these test sentences. Our methods extend those of our earlier work in \cite{Bowman:Potts:Manning:2014}, which introduces an experiment and corresponding artificial dataset to test this ability in two tree models. We adapt that experiment to sequence models by decorating the statements with an explicit bracketing, and we use this design to compare an LSTM sequence model with three tree models, with a focus on what data each model needs in order to generalize well. 

As in \cite{Bowman:Potts:Manning:2014}, we find that standard tree neural networks are able to make the necessary generalizations, with their performance decaying gradually as the structures in the test set grow in size. We additionally find that extending the training set to include larger structures mitigates this decay. Then considering sequence models, we find that a single-layer LSTM is also able to generalize to unseen large structures, but that it does this only when trained on a larger and more complex training set than is needed by the tree models to reach the same generalization performance.

Our results engage with those of \cite{vinyals2014grammar} and \cite{dyer2015transition}, who find that sequence models can learn to recognize syntactic structure in natural language, at least when trained on explicitly syntactic tasks. The simplest model presented in \cite{vinyals2014grammar} uses an LSTM sequence model to encode each sentence as a vector, and then generates a linearized parse (a sequence of brackets and constituent labels) with high accuracy using only the information present in the vector. This shows that the LSTM is able to identify the correct syntactic structures and also hints that it is able to develop a generalizable method for encoding these structures in vectors. However, the massive size of the dataset needed to train that model, 250M tokens, leaves open the possibility that it primarily learns to generate only tree structures that it has already seen, representing them as simple hashes---which would not capture unseen tree structures---rather than as structured objects.
Our experiments, though, show that LSTMs can learn to understand tree structures when given enough data, suggesting that there is no fundamental obstacle to learning this kind of structured representation. We also find, though, that sequence models lag behind tree models across the board, even on training corpora that are quite large relative to the complexity of the underlying grammar, suggesting that tree models can play a valuable role in tasks that require recursive interpretation.

%% file: data.tex
\section{Recursive structure in artificial data}\label{sec:recursion}
\paragraph{Reasoning about entailment} 
The data that we use define a version of the \emph{recognizing textual entailment} task, in which the goal is to determine what kind of logical consequence relation holds between two sentences, drawing on a small fixed vocabulary of relations such as entailment, contradiction, and synonymy. This task is well suited to evaluating neural network models for sentence interpretation: models must develop comprehensive representations of the meanings of each sentence to do well at the task, but the data do not force these representations to take a specific form, allowing the model to learn whatever kind of representations it can use most effectively.

The data we use are labeled with the seven mutually exclusive logical relations of \newcite{maccartney2009extended}, which distinguish entailment in two directions ($\natfor$, $\natrev$), equivalence ($\nateq$), exhaustive and non-exhaustive contradiction ($\natneg$, $\natalt$), and two types of semantic independence ($\natind$, $\natcov$).


\begin{table}[tp]
  \centering\small
    \begin{tabular}[t]{r c l}
      \toprule
      $\plneg p_3$        & $\natneg$ & $p_3$ \\
      $p_3$               & $\natfor$ & $p_3 \plor p_2$ \\
      $(\plneg p_2) \pland p_6  $ & $\natalt$ & $     \plneg ( p_6 \plor ( p_5 \plor p_3 ) ) $\\
 $    p_4 \plor ( \plneg ( ( p_1  \plor p_6 )  \plor p_4 ) ) $ & $ \natfor $ & $  \plneg ( ( ( ( \plneg p_6 ) \plor ( \plneg p_4 ) )  \pland ( \plneg p_5 ) )\pland ( p_6 \pland p_6 ) ) $\\      
      \bottomrule
    \end{tabular}
    \caption{Examples of short to moderate length pairs from the artificial data introduced in \protect\cite{Bowman:Potts:Manning:2014}. We only show the parentheses that are needed to disambiguate the sentences rather than the full binary bracketings that the models use.}\label{tab:plexs}
\end{table}

\paragraph{The artificial language} The language described in \newcite{Bowman:Potts:Manning:2014} (\S4) is designed to highlight the use of recursive structure with minimal additional complexity. Its vocabulary consists only of six unanalyzed word types ($p_1, p_2, p_3, p_4, p_5, p_6$), \word{and}, \word{or}, and \word{not}. Sentences of the language can be straightforwardly interpreted as statements of propositional logic (where the six unanalyzed words types are variable names), and labeled sentence pairs can be interpreted as theorems of that logic. Some example pairs are provided in Table~\ref{tab:plexs}.

Crucially, the language is defined such that any sentence can be embedded under negation or conjunction to create a new sentence, allowing for arbitrary-depth recursion, and such that the scope of negation and conjunction are determined only by bracketing with parentheses (rather than bare word order). The compositional structure of each sentence can thus be an arbitrary tree, and interpreting a sentence correctly requires using that structure.

The data come with parentheses representing a complete binary bracketing. Our models use this information in two ways. For the tree models, the parentheses are not word tokens, but rather are used in the expected way to build the tree. For the sequence model, the parentheses are word tokens with associated learned embeddings. This approach provides the models with equivalent data, so their ability to handle unseen structures can be reasonably compared.

\paragraph{The data}
Our sentence pairs are divided into thirteen bins according to the number of logical connectives (\word{and, or, not}) in the longer of the two sentences in each pair. We test each model on each bin separately (58k total examples, using an 80/20\% train/test split) in order to evaluate how each model's performance depends on the complexity of the sentences. In three experiments, we train our models on the training portions of bins 0--3 (62k examples), 0--4 (90k), and 0--6 (160k), and test on every bin but the trivial bin 0. Capping the size of the training sentences allows us to evaluate how the models interpret the sentences: if a model's performance falls off abruptly above the cutoff, it is reasonable to conclude that it relies heavily on specific sentence structures and cannot generalize to new structures. If a model's performance decays gradually\footnote{Since sentences are fixed-dimensional vectors of fixed-precision floating point numbers, all models will make errors on sentences above some length, and L2 regularization (which helps overall performance) exacerbates this by discouraging the model from using the kind of numerically precise, nonlinearity-saturating functions that generalize best.} with no such abrupt change, then it must have learned a more generally valid interpretation function for the language which respects its recursive structure.

%% file: methods.tex
\section{Testing sentence models on entailment} \label{methods}

We use the architecture depicted in Figure~\ref{fig:model:top}, which builds on the one used in \newcite{Bowman:Potts:Manning:2014}. The model architecture uses two copies of a single sentence model (a tree or sequence model) to encode the premise and hypothesis (left and right side) expressions, and then uses those encodings as the features for a multilayer classifier which predicts one of the seven relations. Since the encodings are computed separately, the sentence models must encode complete representations of the meanings of the two sentences for the downstream model to be able to succeed.

\begin{figure*}[t]
  \centering
  \input{figure1}
  \caption{In our model, two copies of a sentence model---based on either tree (b) or sequence (c) models---encode the two input sentences. A multilayer classifier component (a) then uses the resulting vectors to predict a label that reflects the logical relationship between the two sentences.}
  \label{sample-figure}
\end{figure*}

\paragraph{Classifier}
The classifier component of the model consists of a combining layer which takes the two sentence representations as inputs, followed by two neural network layers, then a softmax classifier.
For the combining layer, we use a neural tensor network (NTN, \cite{chen2013learning}) layer, which sums the output of a plain recursive/recurrent neural network layer with a vector computed using two multiplications with a learned (full rank) third-order tensor parameter:
\begin{gather} 
\label{TreeRNN}
\vec{y}_{\textit{NN}} = \tanh(\mathbf{M} \colvec{2}{\vec{x}^{(l)}}{\vec{x}^{(r)}} + \vec{b}\,) \\
\label{TreeRNTN} 
\vec{y}_{\textit{NTN}} = \vec{y}_{\textit{NN}} + \tanh(\vec{x}^{(l)T} \mathbf{T}^{[1 \ldots n]} \vec{x}^{(r)})
\end{gather} 

Our model is largely identical to the model from \newcite{Bowman:Potts:Manning:2014}, but adds the two additional $\tanh$ NN layers, which we found help performance across the board, and also uses the NTN combination layer when evaluating all four models, rather than just the TreeRNTN model, so as to ensure that the sentence models are compared in as similar a setting as possible.

We only study models that encode entire sentences in fixed length vectors, and we set aside models with attention \cite{bahdanau2014neural}, a technique which gives the downstream model (here, the classifier) the potential to access each input token individually through a soft content addressing system. While attention simplifies the problem of learning complex correspondences between input and output, there is no apparent reason to believe that it should improve or harm a model's ability to track structural information like a given token's position in a tree. As such, we expect our results to reflect the same basic behaviors that would be seen in attention-based models.

\paragraph{Sentence models}
The sentence encoding component of the model transforms the (learned) embeddings of the input words for each sentence into a single vector representing that sentence. We experiment with tree-structured models (Figure~\ref{fig:model:tree}) with TreeRNN (eqn.~\ref{TreeRNN}), TreeRNTN (eqn.~\ref{TreeRNTN}), and TreeLSTM \cite{tai2015improved} activation functions. In addition, we use a sequence model (Figure~\ref{fig:model:seq}) with an LSTM activation function \cite{hochreiter1997long} implemented as in \newcite{zaremba2015recurrent}. In experiments with a simpler non-LSTM RNN sequence model, the model tended to badly underfit the training data, and those results are not included here.

\paragraph{Training} We randomly initialize all embeddings and layer parameters, and train them using minibatch stochastic gradient descent with AdaDelta \cite{zeiler2012adadelta} learning rates. Our objective is the standard negative log likelihood classification objective with L2 regularization (tuned on a separate train/test split). All models were trained for 100 epochs, after which all had largely converged without significantly declining from their peak performances.

%% file: figure1.tex
\begin{subfigure}[t]{0.45\textwidth}
\centering
\scalebox{0.65}{
 \begin{tikzpicture}
    \def\dx{21pt}
    \def\dy{27pt}

    \tikzstyle{label}=[text width=40mm,align=center,text height=2mm]    
    \tikzstyle{softmax}=[fill=red!40,text width=40mm,align=center,text height=2mm]
    \tikzstyle{preclass}=[fill=orange!40,text width=50mm,align=center,text height=2mm]
    \tikzstyle{e}=[fill=cyan!40,text width=26mm,align=center,text height=2mm]
    \tikzstyle{m}=[draw=black,text width=38mm,align=center,text height=2mm]    
    
    \node[softmax]  (softmax) at (0*\dx,6*\dy) {7-way softmax classifier};
    \node[preclass,fill=orange!40]  (pc3) at (0*\dx,5*\dy) {100d $\tanh$ NN layer};
    \node[preclass,fill=yellow!40]  (pc2) at (0*\dx,4*\dy) {100d $\tanh$ NN layer};
    \node[preclass,fill=green!40]  (pc1) at (0*\dx,3*\dy) {100d $\tanh$ NTN layer};
    \node[e]  (pe) at (-3*\dx,1.75*\dy) {50d premise};
    \node[e]  (he) at (3*\dx,1.75*\dy) {50d hypothesis};
    \node[m]  (pem) at (-3*\dx,0.5*\dy) {sentence model\\ with premise input};
    \node[m]  (hem) at (3*\dx,0.5*\dy) {sentence model\\ with hypothesis input};    
    
    \pgfsetarrowsend{latex}
    \tikzstyle{fwd} = [draw=black, line width=1pt]

          \draw [fwd] (pc3) -- (softmax);
          \draw [fwd] (pc2) -- (pc3);
          \draw [fwd] (pc1) -- (pc2);
          \draw [fwd] (pe) -- (pc1);
          \draw [fwd] (he) -- (pc1);
          \draw [fwd] (hem) -- (he);
          \draw [fwd] (pem) -- (pe);

  \end{tikzpicture}}
  
 \caption{The general architecture shared across models.}\label{fig:model:top}
  
\end{subfigure}\\
\begin{subfigure}[t]{0.45\textwidth}
  \centering
\scalebox{0.65}{
 \begin{tikzpicture}
    \def\dx{21pt}
    \def\dy{27pt}

    \tikzstyle{word}=[fill=purple!40,text width=16mm,text height=2mm,align=center]
    \tikzstyle{node}=[fill=blue!40,text width=16mm,text height=2mm,align=center]
    \tikzstyle{empty}=[fill=blue!0,text width=8mm,text height=2mm,align=center]

    \node[empty]  (null) at (0*\dx,7*\dy) {...};
    \node[node]  (aorb) at (0*\dx,6*\dy) {a or b};
    \node[word]  (a) at (-2.5*\dx,5*\dy) {a};
    \node[node]  (orb) at (2.5*\dx,5*\dy) {or b};
    \node[word]  (or) at (0*\dx,4*\dy) {or};
    \node[word]  (b) at (5*\dx,4*\dy) {b};

    \pgfsetarrowsend{latex}
    \tikzstyle{fwd} = [draw=black, line width=1pt]

          \draw [fwd] (or) -- (orb);
          \draw [fwd] (b) -- (orb);
          \draw [fwd] (a) -- (aorb);
          \draw [fwd] (orb) -- (aorb);
          \draw [fwd] (aorb) -- (null);
  \end{tikzpicture}}
  
     \caption{The architecture for the tree-structured sentence models. Terminal nodes are learned embeddings and nonterminal nodes are NN, NTN, or TreeLSTM layers.}\label{fig:model:tree}
  
  \end{subfigure}\qquad
\begin{subfigure}[t]{0.45\textwidth}
  \centering
\scalebox{0.65}{
 \begin{tikzpicture}
    \def\dx{35pt}
    \def\dy{27pt}

    \tikzstyle{word}=[fill=purple!40,text width=16mm,text height=2mm,align=center]
    \tikzstyle{node}=[fill=blue!40,text width=16mm,text height=2mm,align=center]
    \tikzstyle{empty}=[fill=blue!0,text width=8mm,text height=2mm,align=center]

    \node[word]  (a) at (-3*\dx,1*\dy) {a};
    \node[node]  (aN) at (-3*\dx,2*\dy) {a};
    
    \node[word]  (or) at (-1*\dx,1*\dy) {or};
    \node[node]  (orN) at (-1*\dx,2*\dy) {a or};
    
    \node[word]  (b) at (1*\dx,1*\dy) {b};
    \node[node]  (bN) at (1*\dx,2*\dy) {a or b}; 
    
    \node[empty]  (nullN) at (2.75*\dx,2*\dy) {...}; 
    
    \pgfsetarrowsend{latex}
    \tikzstyle{fwd} = [draw=black, line width=1pt]

          \draw [fwd] (or) -- (orN);
          \draw [fwd] (b) -- (bN);
          \draw [fwd] (a) -- (aN);
          \draw [fwd] (aN) -- (orN);
          \draw [fwd] (orN) -- (bN);
          \draw [fwd] (bN) -- (nullN);
          
  \end{tikzpicture}}
  
   \caption{The architecture for the sequence sentence model. Nodes in the lower row are learned embeddings and nodes in the upper row are LSTM layers.}\label{fig:model:seq}
  
    \end{subfigure}

%% file: discussion.tex
\section{Results and discussion}\label{sec:discussion}

\begin{figure*}[t]
  \centering
  \begin{subfigure}[t]{0.04\textwidth}
      \includegraphics[height=1.25in]{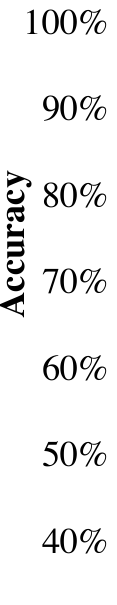}
\end{subfigure}
\begin{subfigure}[t]{0.24\textwidth}
  \includegraphics[height=1.25in]{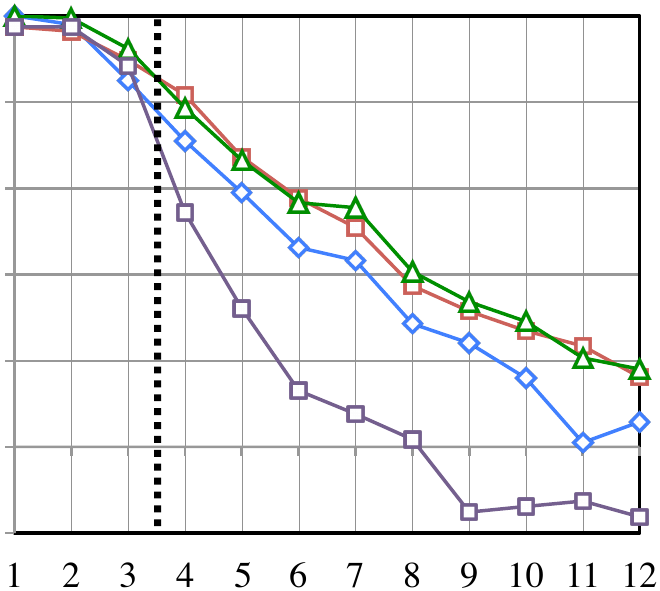}
  \caption{Training on sz. $\le$3.}
  \end{subfigure}~~~
\begin{subfigure}[t]{0.24\textwidth}
    \includegraphics[height=1.25in]{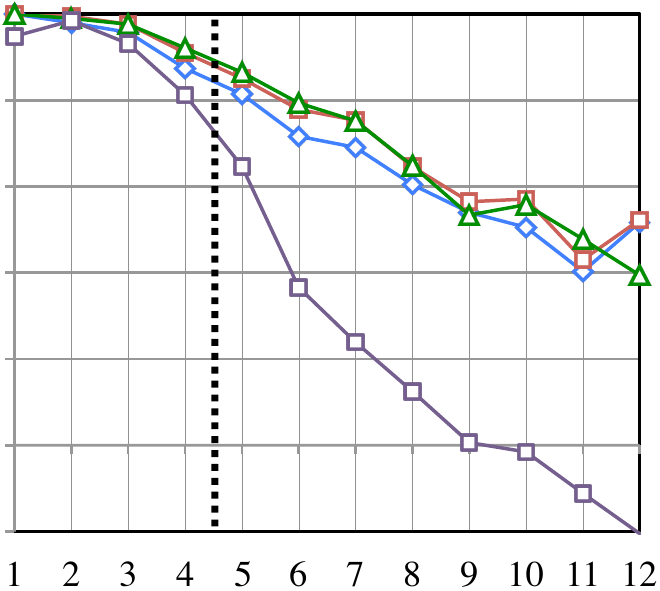}
  \caption{Training on sz. $\le$4.}
  \end{subfigure}~~~
\begin{subfigure}[t]{0.24\textwidth}
      \includegraphics[height=1.25in]{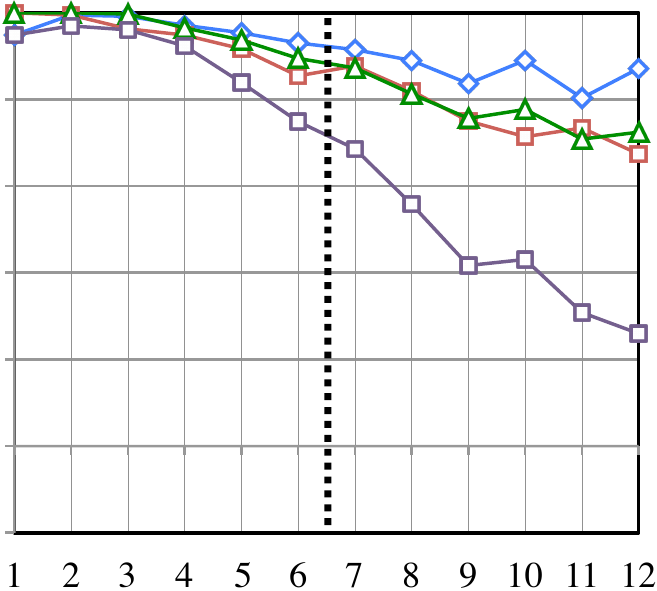}
  \caption{Training on sz. $\le$6.}
\end{subfigure}~~
\begin{subfigure}[t]{0.08\textwidth}
      \includegraphics[height=1.25in]{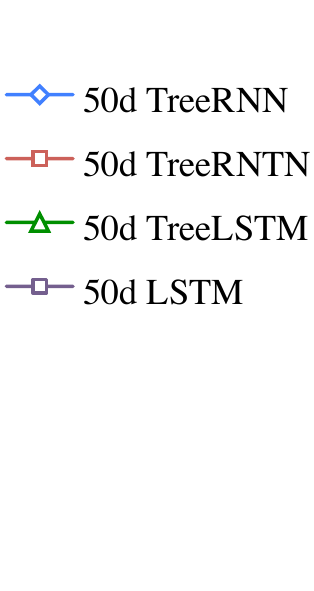}
\end{subfigure}
  \caption{Test accuracy on three experiments with increasingly rich training sets. The horizontal axis on each graph divides the test set expression pairs into bins by the number of logical operators in the more complex of the two expressions in the pair. The dotted line shows the size of the largest examples in the training set in each experiment.}
  \label{prop-results} 
\end{figure*}

The results are shown in Figure~\ref{prop-results}. 
The tree models fit the training data well, reaching 98.9, 98.8, and 98.4\% overall accuracy respectively in the $\le$6 setting, with the LSTM underfitting slightly at 94.8\%. 
In that setting, all models generalized well to structures of familiar length, with the tree models all surpassing 97\% on examples in bin 4, and the LSTM reaching 94.8\%.
On the longer test sentences, the tree models decay smoothly in performance across the board, while the LSTM decays more quickly and more abruptly, with a striking difference in the $\le$4 setting, where LSTM performance falls 10\% from bin 4 to bin 5, compared to 4.4\% for the next worse model. However, the LSTM improves considerably with more ample training data in the $\le$6 condition, showing only a 3\% drop and generalization results better than the best model's in the $\le$3 setting.

All four models robustly beat the simple baselines reported in \newcite{Bowman:Potts:Manning:2014}: the most frequent class occurs just over 50\% of the time and a neural bag of words model does reasonably on the shortest examples but falls below 60\% by bin~4.

\begin{figure}[t]
  \centering
      \includegraphics[height=1.1in]{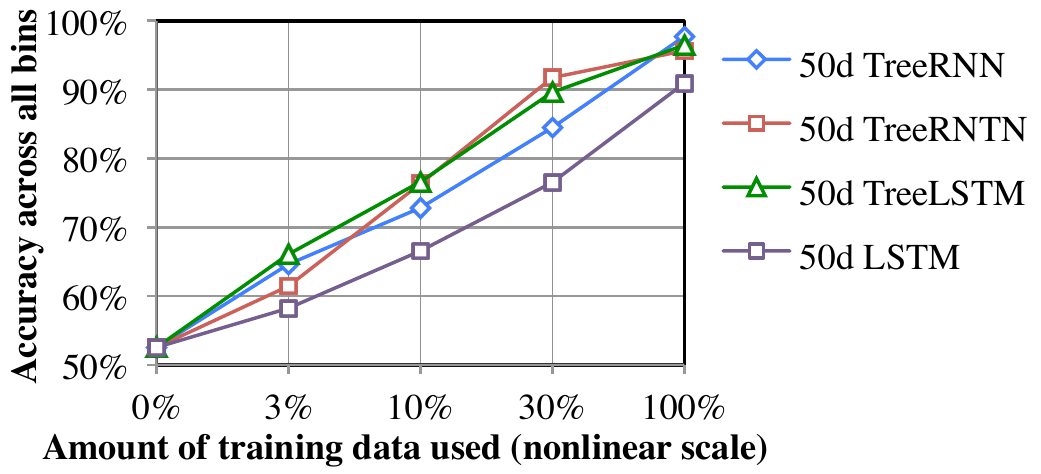}
  \caption{Learning curve for the $\le$6 experiment.}
  \label{fig:lc} 
\end{figure}

The learning curve (Figure~\ref{fig:lc}) suggests that additional data is unlikely to change these basic results. The LSTM lags behind the tree models across the curve, but appears to gain accuracy at a similar rate.

\section{Conclusion}

We find that all four models are able to effectively exploit a recursively defined language to interpret sentences with complex unseen structures.
We find that tree models' biases allow them to do this with greater efficiency, outperforming sequence-based models substantially in every experiment. However, our sequence model is nonetheless able to generalize smoothly from seen sentence structures to unseen ones, showing that its lack of explicit recursive structure does not prevent it from recognizing recursive structure in our artificial language.

We interpret these results as evidence that both tree and sequence architectures can play valuable roles in the construction of sentence models over data with recursive syntactic structure. Tree architectures provide an explicit bias that makes it possible to efficiently learn to compositional interpretation, which is difficult for sequence models. Sequence models, on the other hand, lack this bias, but have other advantages. Since they use a consistent graph structure across examples, it is easy to accelerate minibatch training in ways that yield substantially faster training times than are possible with tree models, especially with GPUs. In addition, when sequence models integrate each word into a partial sentence representation, they have access to the entire sentence representation up to that point, which may provide valuable cues for the resolution of lexical ambiguity, which is not present in our artificial language, but is a serious concern in natural language text.

Finally, we suggest that, because of the well-supported linguistic claim that the kind of recursive structure that we study here is key to the understanding of real natural languages, there is likely to be value in developing sequence models that can more efficiently exploit this structure without fully sacrificing the flexibility that makes them succeed.